%% file: 0_main.tex
\documentclass[letterpaper, 10 pt, conference]{ieeeconf}
\IEEEoverridecommandlockouts
\usepackage{cite}
\usepackage{amsmath,amssymb,amsfonts}
\usepackage{algorithmic}
\usepackage{graphicx}
\usepackage{textcomp}
\usepackage{xcolor}
\usepackage{balance}
\usepackage[capitalize]{cleveref}
\usepackage{adjustbox}

\def\BibTeX{{\rm B\kern-.05em{\sc i\kern-.025em b}\kern-.08em
    T\kern-.1667em\lower.7ex\hbox{E}\kern-.125emX}}

\title{\LARGE \bf Two-Finger Soft Gripper Force Modulation via Kinesthetic Feedback}

\author{Stephanie O. Herrera, Tae Myung Huh, Dejan Milutinović
\thanks{All authors are with Dept. of Electrical and Computing Engineering, University of California Santa Cruz, Santa Cruz, CA, USA. email: {\tt\small soherrer@ucsc.edu}}%
}

\begin{document}
\maketitle

\thispagestyle{empty}
\pagestyle{empty}

\global\csname @topnum\endcsname 0
\global\csname @botnum\endcsname 0

\begin{abstract}
We investigate a method to modulate contact forces between the soft fingers
of a two-finger gripper and an object, without relying on tactile sensors. 
This work is a follow-up to our previous results on contact detection. Here, 
our hypothesis is that once the contact between a finger and an object is 
detected, a controller that keeps a desired difference between the finger 
bending measurement and its bending at the moment of contact is sufficient 
to maintain and modulate the contact force. This approach can be simultaneously 
applied to both fingers while getting in contact with a single object. 
We successfully tested the hypothesis, and characterized the contact and peak pull-out 
force magnitude vs. the desired difference expressed by a multiplicative 
factor. All of the results are performed on a real physical device.
\end{abstract}

\input{1_Intro}

\input{2_ProblemFormulation}

\input{3_Controller}

\input{4_ExperimeintResult}
\input{5_Discussion}

\bibliographystyle{IEEEtran}

\balance  

\bibliography{IEEEabrv,Biblio}   
\vspace{\baselineskip}

\end{document}

%% file: 1_Intro.tex
\section{Introduction}

Soft robotic grippers hold significant potential for tasks beyond simple pick-and-place operations, enabling more sophisticated dexterous manipulation. While demonstrations and models of such manipulation have been presented in the literature\cite{9761831,9134855,9495192}, most do not incorporate contact force measurements or force modulation, relying instead on the inherent compliance of the gripper in open-loop systems. However, to achieve more advanced functions, such as in-hand manipulation in unstructured environments with unknown object properties, it becomes essential to understand and control the forces involved, enhancing precision and effectiveness.

Tactile sensors, even in soft grippers, have been the primary method for detecting contact and measuring contact forces \cite{deng2022design,luo2022digital,qu2023recent}. Although tactile sensors provide direct force measurements, their effectiveness can be limited by the complexity of integration and the wear and tear of contact surfaces over time, which may compromise consistent and reliable sensing.

\begin{figure}[tp!]
\centerline{\includegraphics[scale=0.3]{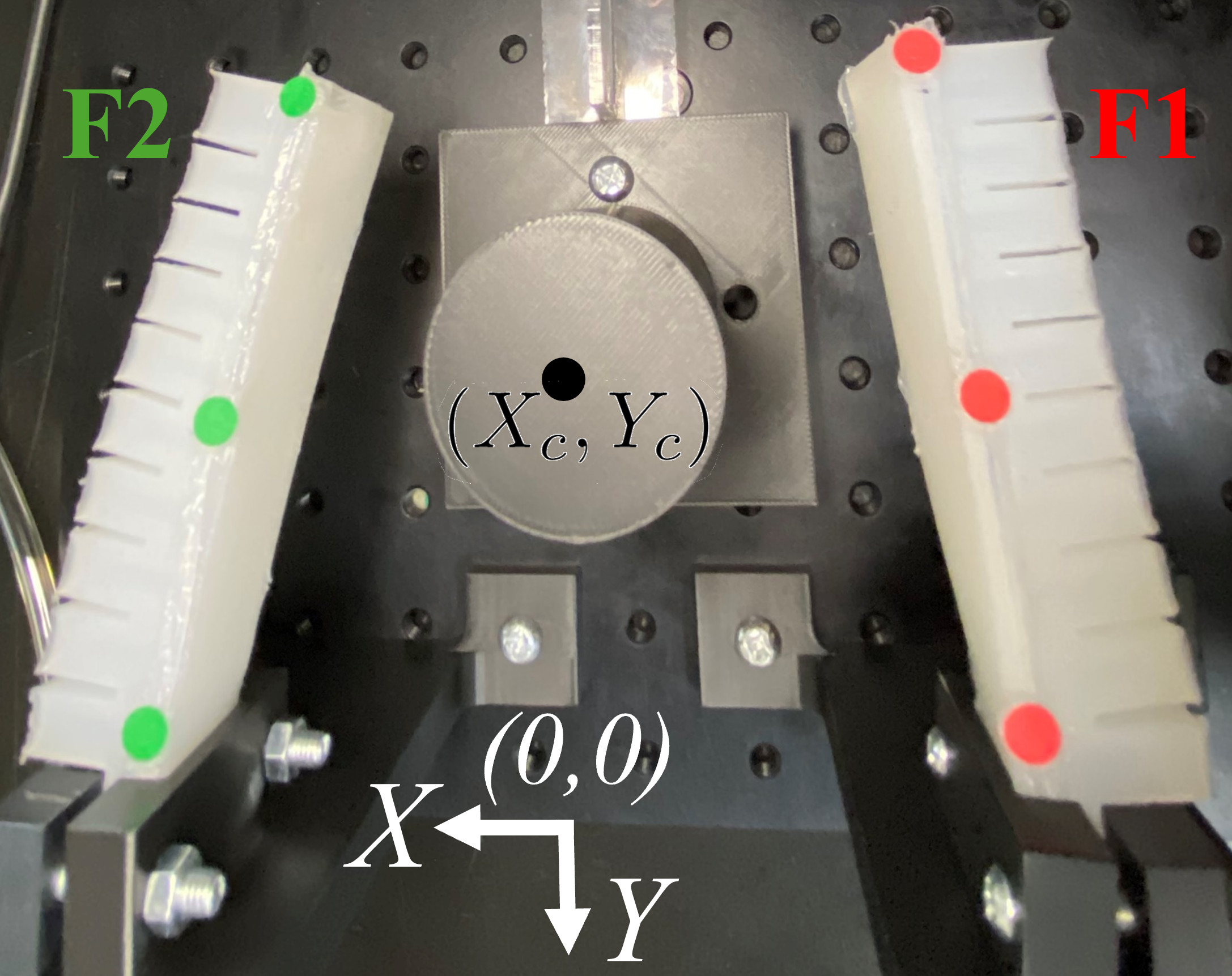}}
\caption{Two-finger soft gripper experiment setup: Fingers F1 and 
F2 are actuated by feedback control using their bending measurements 
and outputting a control variable for the corresponding finger inflation 
pressure. The cylindrical object positioned at $(X_c,Y_c)$ is mounted 
on a force sensor which we use to measure contact forces  
depicted along the $X$ and $Y$ directions. The finger design is based on \cite{6696549}.}
\label{fig1}
\end{figure}

Motivated by these limitations, we propose a feedback control-based kinesthetic sensing approach for soft grippers, where contact events are detected by monitoring changes in the reference tracking error, as demonstrated in our previous work\cite{8815142,15730409,10155854,JINTBoivin2022}. Kinesthetic sensing in humans refers to the awareness of body movement and sensory feedback from receptors within muscles, tendons, and joints \cite{Dahiya2010}. Similarly, bending sensors—either embedded within the gripper fingers or used remotely—can detect interaction forces by comparing desired finger bending with 
actual bending measurements. In our earlier studies, we explored this kinesthetic sense of touch using a single robotic soft-finger embedded with a resistive bending sensor, successfully detecting contact between the finger and an object\cite{JINTBoivin2022} by identifying the moment when the 
curvature reference tracking error increased due to contact force with the object. Recently 
published work\cite{retrofit}, based on fluidic input sensing (rather than bending), has also 
demonstrated similar contact detection between a soft finger and the environment. However, 
these approaches have yet to modulate the contact force during interaction.

Following up on our previous work, we adopt here 
the previously used reference tracking feedback controller structure and 
design the procedure from \cite{8815142} to investigate the use of feedback 
control to modulate contact forces between two soft fingers and an object.
Therefore, our paper contributions are:
\begin{itemize}
\item validation of the reference tracking controller from \cite{8815142} 
in our setup with different soft fingers and bending measurements;
\item feedback control for modulating contact forces between soft 
fingers and an object using only soft finger bending measurements;
\item characterization of the feedback control-enabled contact and 
peak pull-out forces between two soft fingers and an object.
\end{itemize}

This paper is organized as follows. Section II outlines the problem 
statement concerning the modulation of contact forces. In Section III, we present preliminary 
findings of the bending reference tracking controller, which are part of our validation of the reference 
tracking controller. Section IV describes the force-modulating controller and its design. 
Section V reports characterization experiments, and Section VI provides conclusions based on our results.

%% file: 2_ProblemFormulation.tex
\section{Problem Formulation}
\begin{figure}[tp!]
\centerline{\includegraphics[width=\linewidth]{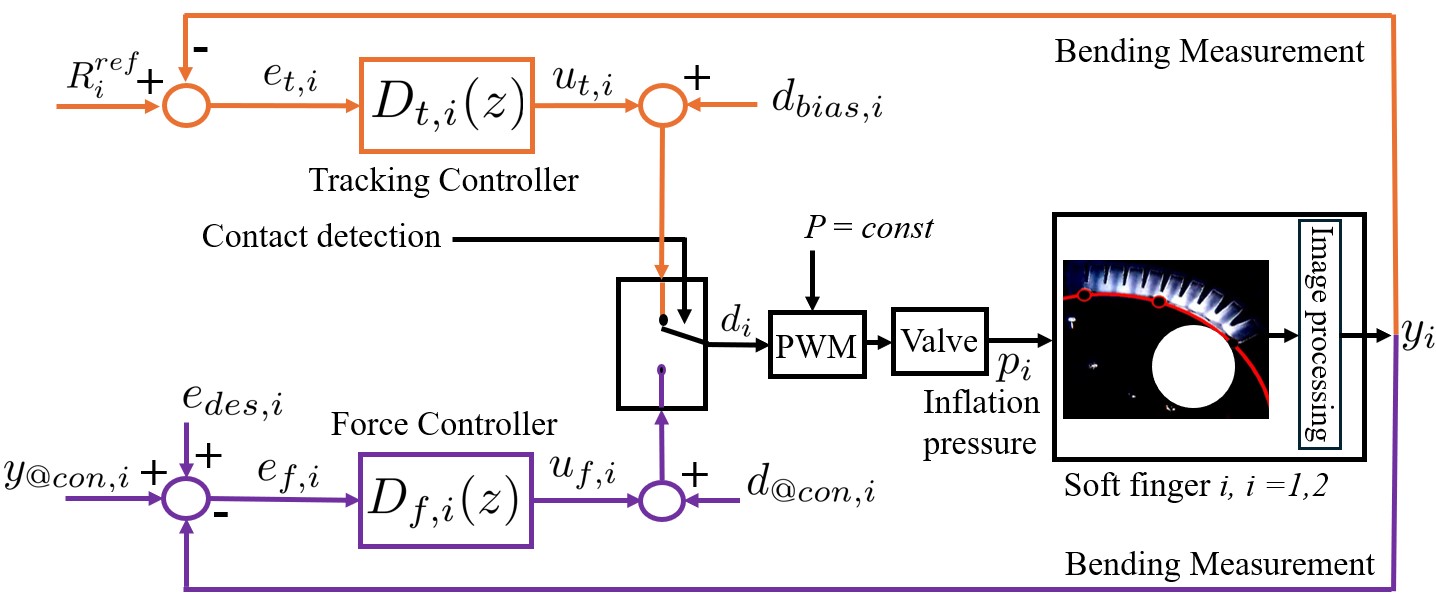}}
\caption{
The soft finger feedback control: $R_{i}^{ref}$ is the bending reference for 
the bending measurement $y_i$ of the control loop with the tracking controller 
$D_{t,i}(z)$ (orange). The contact detection, which is based on magnitudes of 
$e_{t,i}$ and $d_i$, switches to the force modulating control with the 
controller $D_{f,i}(z)$ (purple). The signals $y_{@con,i}$ and $d_{@con,i}$ 
are bending and duty cycle at the time of contact, $d_{bias,i}$ compensates
for the dead zone in actuation, and $P$ depicts the constant air pressure source.}
\label{fig2}
\end{figure}

Figure~\ref{fig2} depicts two feedback control loops for the control of a single 
finger in Fig.~\ref{fig1}. One of the loops includes the controller $D_{t,i}(z)$ and 
the other the controller $D_{f,i}(z)$. Both controllers use the same 
bending-related measurement $y_i$ and output a duty cycle $d_i$, which dictates 
the finger inflation pressure $p_i$. At any time, only one of the feedback control 
loops is active.

The controller $D_{t,i}(z)$ is part of the \emph{reference tracking} control loop which 
enables contact detection between the finger and an object. The controller was designed 
and demonstrated in \cite{JINTBoivin2022} using resistive bending sensitive sensors 
mounted onto the dorsal side of the finger. In \cite{JINTBoivin2022}, it was stated that 
the controller design process could be applied to other types of bending-sensitive sensors 
and measurements. Therefore, our first problem (P1) \emph{is to validate the use of 
the reference tracking controller on the two-finger gripper in Fig.~\ref{fig1}}.

After contact is detected, there is a switch from the reference tracking control 
to \emph{force modulation control} with the controller $D_{f,i}(z)$. 
We propose to design the controller $D_{f,i}(z)$ that, for the desired 
error value $e_{des,i}=0$, results in the controller input 
$e_{f,i}$ convergence to 0, and the bending measurement $y_{i}$ convergence to the 
value of the bending measurement at 
the moment of contact detection $y_{@con,i}$. Under these conditions, 
the contact of the finger with an object is a gentle or light one \cite{JINTBoivin2022}.

We hypothesize that if we set $e_{des,i}>0$, it will induce an increase in the 
inflation pressure, i.e., the finger bending further and a larger contact force. 
Therefore, our second problem (P2) \emph{is to test the hypothesis about force 
modulation using $e_{des,i}$} and the controller $D_{f,i}(z)$. 

Due to the elasticity of the soft finger body, we expect that the relation between
$e_{des,i}$ and the contact force is linear, or at least monotonously increasing 
and similar for both fingers. For this reason, our goal here (P3) \emph{is to 
characterize the relation between $e_{des,i}$ vs. contact forces and two-finger 
pull-out forces.}

%% file: 3_Controller.tex
\section{Technical Preliminaries - Reference Tracking Controller and Contact Detection}
 
\begin{figure}[t!]
\centerline{\includegraphics[width=0.7\linewidth]{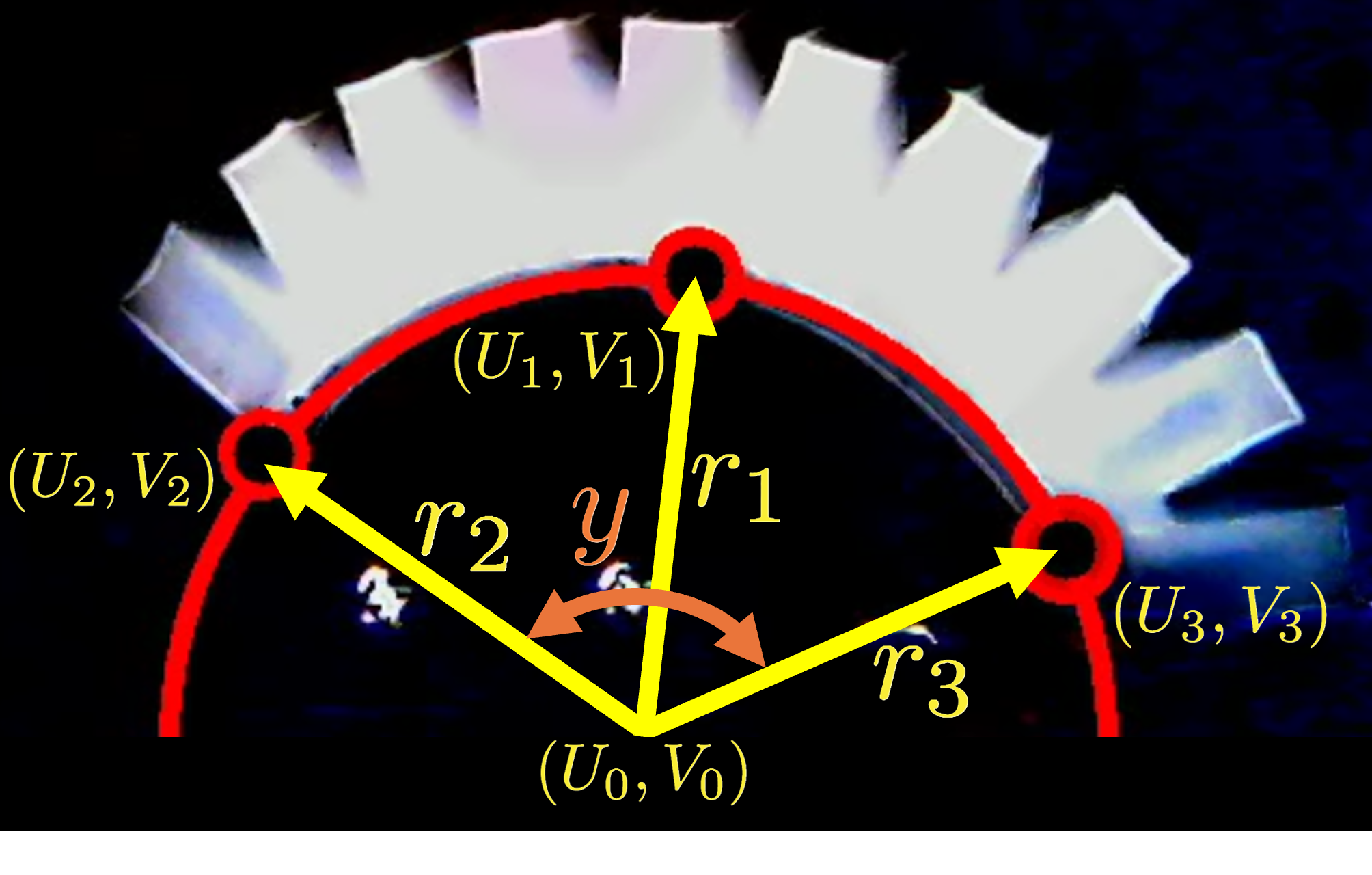}}
\caption{
Bending measurements $y$: $(U_k,V_k)$, $k=1,2,3$ are pixel 
coordinates of the centers of the circular markers, $(U_0,V_0)$ 
is the center of a circle fitted to the three markers. Vector magnitudes $|r_k|$, $k=1,2,3$ are identically equal.
}
\label{fig:bending}
\end{figure}

Both feedback control loops depicted in Fig.~\ref{fig2} 
depend on the finger-related bending measurement. The measurement 
is computed from the camera video frame-detected color markers 
as shown in Fig.~\ref{fig:bending}. The figure also shows that 
frame-detected markers are not labeled in any particular order, 
which is a consequence of using OpenCV functions. For this reason, 
the bending $y$ is computed as
\begin{equation}
y = \max_{k,j, k\ne j} \left\{ {\rm arccos} \frac{r_k^T r_j}{|r_k||r_j|} \right\},\ \ \ k,j=1,2,3,
\end{equation}
where $r_k=[U_k-U_0, V_k-V_0]^T$, which is a column corresponding
to the vector ${\rm r}_k$ from Fig.~\ref{fig:bending} which starts at $(U_0, V_0)$ 
and ends at $(U_k, V_k)$  pixel coordinates. $|r_k|$ is the vector magnitude and 
the superscript ${}^T$ denotes a matrix transpose. In the sequel, we use $y_i$, $i=1,2$ 
for the bending $y$ measurement computed for each finger and express it in 
degrees.

\emph{Reference Tracking Controller}: Here, we follow a process of the controller design as described in \cite{JINTBoivin2022}. In this process, we use a pseudo-random sequence of duty cycles ($d$) and 
the corresponding bending measurements ($y$), both sampled with a sample time of $T_s=0.01$ $sec$. 
Using system identification, we estimate a 2nd order ARX 
model (AutoRegressive with eXogenous inputs model) 
from $d$ to $y$, which has a z-domain form 
\begin{equation} \label{eq:armax2}
A(z)y(z) = B(z)d(z) + e_m(z),
\end{equation}
where $y(z)$, $d(z)$ and $e_m(z)$ are z-domain images of the output, 
input and modeling errors, and  $A(z)$ and $B(z)$ are polynomials   
\begin{equation} 
\begin{aligned}
A(z) &= z^2+a_1z+a_2=z^2 - 0.7655z + 0.03624, \\
B(z) &= b_1z+b_2=0.04074z + 0.3601.
\end{aligned} \label{eq:armax}
\end{equation}
This yields that the z-domain transfer function $G(z)$ of the soft 
finger is 
\begin{equation}
 G(z) = \frac{B(z)}{A(z)}.
\end{equation}
The analysis in \cite{JINTBoivin2022} shows that the 
ramp-signal tracking error $e_{t,i}$ convergence to 0, i.e., 
zero-error tracking of a ramp reference, can be achieved with 
a controller that has the following z-domain transfer function
\begin{equation*} \label{eq:sample2} 
D_{t,i}(z) = K_{t} \frac{z - z_{t}}{(z-1)^2}.
 \tag{2}
\end{equation*}
We used the discrete-time root locus analysis in MATLAB's rltool 
to tune the controller $D_{t,i}(z)$ parameter values to $z_{t} = 0.97$ and 
$K_{t} =0.066$.
\begin{figure}[b!]
\centerline{\includegraphics[scale=0.38]{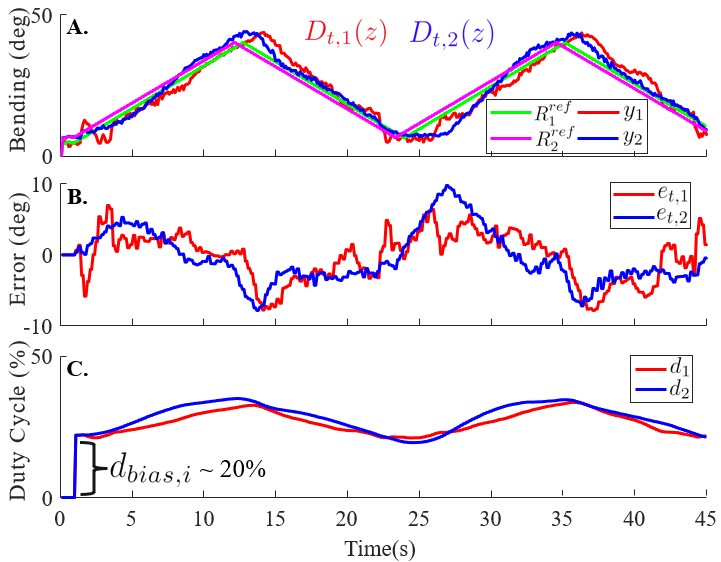}}
\caption{{Bending reference tracking:  (A) Triangular reference inputs $R^{ref}_1$(green) 
for Finger~1 (F1) and $R^{ref}_2$(magenta) for Finger~2 (F2). The bending measurements 
$y_1$(red) for F1 and $y_2$(blue) for F2. (B) The tracking errors $e_{t,1}$(red) for F1 and 
$e_{t,2}$(blue) for F2 . (c) The duty cycle $d_1$(red) for F1 and $d_2$(blue) for F2 control outputs. 
During our experiments, we found that a dead zone of control action was approximately $20$\%.}}
\label{figTracking}
\end{figure}

The use of the tuned feedback controller is illustrated by the experimental 
results in Fig.~\ref{figTracking}. Figure~\ref{figTracking}A
shows bending measurement data for both fingers (red and blue lines) 
when the triangular reference signals (green and magenta lines), slightly 
shifted in time, were used as reference signals for the feedback control 
of the fingers in Fig.~\ref{fig1}. In the experiment, both fingers were 
free to move without any contact among them or with another object. 
As a result of the reference tracking, the fingers periodically flexed 
and extended. The tracking errors $e_{t,i}$ are largest after the change in reference 
slopes, which creates error jumps as seen in Fig.~\ref{figTracking}B. The same plot also shows that during the periods of positive 
or negative slopes, the measurements converge to the corresponding references.
This behavior is expected since the controller $D_{t,i}(z)$ is designed as a ramp reference tracking controller. 

Figure~\ref{figTracking}B also shows the type of data used to estimate 
the empirical cumulative distribution functions (CDFs) of the tracking error for both fingers. We found that $80\%$ of recorded errors are below $e_{tr,1}=3.88$ and $e_{tr,2}=4.71$ for Finger 1 and Finger 2, respectively. We use these values as threshold values for 
contact detection since the largest errors appear when the slope of the reference changes.
However, the reference signal slope change will never happen during the grasping of an object since the fingers are going to be in contact with the object during the first positive slope of the reference.

Figure~\ref{figTracking}C shows the output of the controller $d_i=u_{t,i}+d_{bias,i}$, which is the duty cycle that produces the inflation pressure $p_i$ from the constant input pressure source $P$. In the process of tuning 
the controller, we found that the valve through which we generated $p_i$ had a dead zone of about $20\%$ of the duty cycle, which is the value $d_{bias,i}$ added to the controller $D_{t,i}(z)$ output.

\emph{Contact Detection}: Figure~\ref{figTracking} shows the plots corresponding to freely moving 
soft-fingers. However, if the motion of the fingers is impeded by contact with an object, the 
reference tracking error increases while the duty cycle is well above the dead zone duty cycle 
of $20\%$. At the time point indicated by the red (blue) dashed line in Fig.~\ref{gentle}, the duty cycle is 
at $30\%$ and Finger 1 (Finger~2) is above its error threshold $e_{tr,1}$ ($e_{tr,2}$), so we can 
conclude that Finger 1 (Finger~2) is in contact with the object. This triggers the switch from 
the tracking controller, $D_{t,i}(z)$, to the force controller, $D_{f,i}(z)$, for Finger 1 (Finger~2). 
While the contact detection can be further improved or formulated differently, it is not in the 
focus of our work here. The focus is on the method of force modulation after the contact detection 
when the controller $D_{f,i}(z)$ is active, which we discuss in the following section. 

\section{Force Modulation Controller}

The role of the force modulating controller $D_{f,i}(z)$ in our approach 
is not only to maintain contact between the fingers and an object, but also to allow 
us to modulate the contact force. Since we memorize the bending measurement 
$y_{@con,1}$, $y_{@con,2}$ and the inflation pressure $p_{@con,1}$, $p_{@con,2}$ 
at the moment of contact, with $e_{des,i}=0$ and using $D_{f,i}(z)$ providing the 
convergence of $e_{f,i}$ to 0, we can maintain the contact between the 
finger and an object with a minimal force (light-touch). A controller 
that provides this for the 2nd order model (\ref{eq:armax2})-(\ref{eq:armax})
has the form 
\begin{equation} \label{eq:sample3}
D_{f,i}(z)=K_f\frac{z-z_f}{z-1},
\end{equation}
where $K_{f}$ is the controller gain, $z_{f}$ is the real zero introduced 
by the controller and in our design the term $z-1$ ensures the zero steady 
state error for tracking a step reference signal. The latter can be verified 
by evaluating $e_{f,i}(\infty)=lim_{t \rightarrow \infty} e_{f,i}(t)$ 
using the finite value theorem in the z-domain as
\begin{eqnarray}
 e_{f,i}(\infty) =& \lim_{z \rightarrow 1} (z-1) \frac{1}{1+ D_{f,i}(z)\frac{B(z)}{A(z)}}\frac{zR}{z-1} \\
      =& \lim_{z \rightarrow 1} \frac{R}{1+K_f \frac{1-z_f}{z-1}\frac{b_1+b_2}{1+a_1+a_2}} = 0,
\end{eqnarray}
where $R$ is the magnitude of a step reference signal at $e_{des,i}$.
Our hypothesis from $P2$ is that using the same controller $D_{f,i}(z)$ 
and $e_{des,i}>0$, we can increase the contact force.

In all our experiments, the controller $D_{f,i}(z)$ has the parameters $K_f = 0.21$ and $z_f = 0.44$. 
We discuss the parameter $K_f$ and $z_f$ selection process after showing the results that our hypothesis from $P2$ is correct. 

We first performed experiments with the force sensor-mounted  cylindrical object from 
Fig.~\ref{fig1}, which is positioned between the two fingers at $(0,15)$ $ cm$. 
Consequently, we measured forces that the fingers applied to the object in the $X$ and 
$Y$ directions and denoted them as $F_X$ and $F_Y$, respectively. In our experiments,
the force $F_X$ is the difference between Finger 1 and Finger 2 forces on the object in 
the $X$ direction while the force $F_Y$ is a sum of Finger 1 and Finger 2 forces in the $Y$ 
direction. The fingers were first controlled by their corresponding tracking 
controllers $D_{t,i}(z)$ with ramp references of the same slope that were
initiated at different times (first Finger 1 and then Finger 2). Finger 1 was 
the first to get in contact with the object, switching from the $D_{t,i}(z)$  to 
$D_{f,i}(z)$ controller.

\begin{figure}[tp!]
\centerline{\includegraphics[scale=0.4]{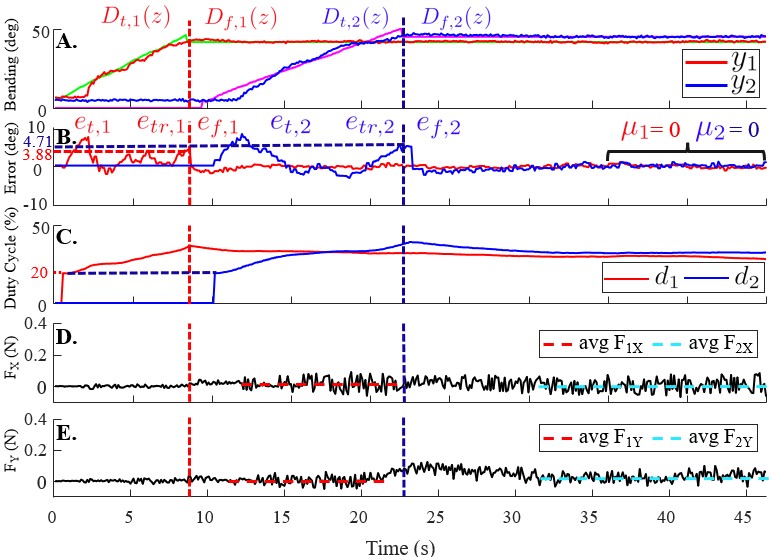}}
\caption{Two-finger gentle force contact: (A) the solid green line is 
$R^{ref}_1$ for the $D_{t,1}(z)$ controller and $y_{con,1}$ for the $D_{f,1}(z)$ controller . The 
solid magenta line is $R^{ref}_2$ for the $D_{t,2}(z)$ controller and $y_{con,2}$ 
for the $D_{f,2}(z)$ controller. (B) Plots of controller tracking errors. The switch from 
 $D_{t,1}(z)$ to $D_{f,1}(z)$ is when $e_{t,1}\ge e_{tr,1}=3.88$ and the switch from 
 $D_{t,2}(z)$ to $D_{f,2}(z)$ is when $e_{t,2}\ge e_{tr,2}=4.71$. (C) Duty cycles 
 $d_1$(red) and $d_2$(blue) for the inflation pressure of $F_1$ and $F_2$, respectively. 
 (D) and (E) are plots of $X$ and $Y$ force components for $e_{des,1}=e_{des,2}=0$ on 
 the cylindrical object of a radius $r=2 cm$ and positioned at the location 
 ($X_c, Y_c$) = (0, 15)\,cm.}\label{gentle}
\end{figure}

\emph{Light touch}: The experiment results are plotted in 
Fig.~\ref{gentle}. At the moment of Finger 1 contact detection, denoted as dashed red lines, we set $R^{ref}_1=y_{@con,1}$ and $e_{des,1}=0$. 
After that time point  ($9$ $s$), we see in Fig.~\ref{gentle}B that the controller 
error $e_{f,1}$ is close to 0 and Fig.~\ref{gentle}D and  
Fig.~\ref{gentle}E show a gentle contact force of 0.006 N. 
At the moment of Finger 2 contact detection  ($23$ $s$), denoted by dashed blue lines, 
we set $R^{ref}_2=y_{@con,2}$ and $e_{des,2}=0$. Following this, we 
see in Fig.~\ref{gentle}B that the corresponding error $e_{f,2}$ 
is also close to 0 and that the measured forces $F_X$ and $F_Y$ plotted 
in Fig.~\ref{gentle}D and Fig.~\ref{gentle}E are still small (0.03 N). This is all expected from the experiment with $e_{des,i}=0$, $i=1,2$. The magnitude 
of the light touch forces as a function of $e_{des,i}$, including 
$e_{des,i}=0$, is examined in Section~\ref{secVB}.

\begin{figure}[b!]
\medskip
\centerline{\includegraphics[width=0.91\linewidth]{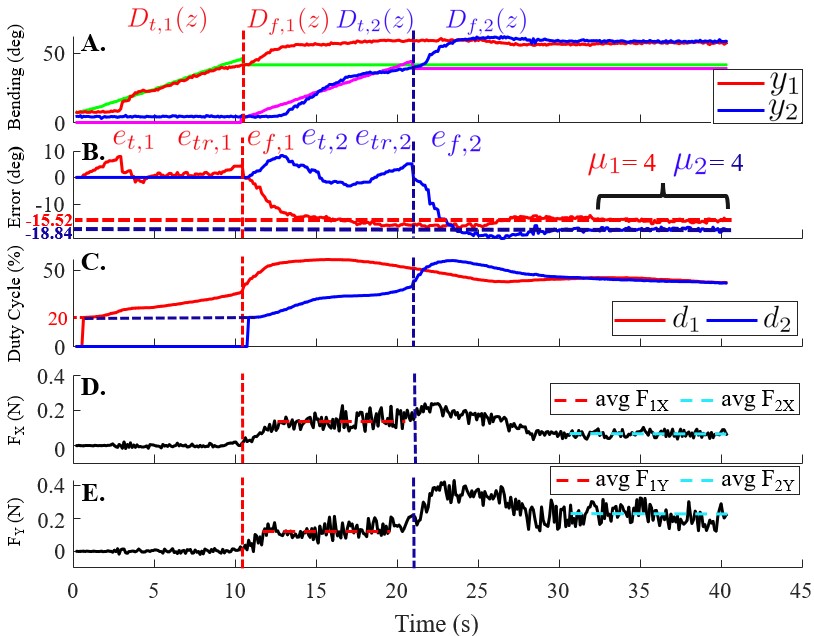}}
\caption{Two-finger force modulation: 
(A) The solid green line is $R^{ref}_1$ for the $D_{t,1}(z)$ controller and $y_{con,1}$ for the $D_{f,1}(z)$ controller. The
solid magenta line is $R^{ref}_2$ for the $D_{t,2}(z)$ controller and $y_{con,2}$ 
for the $D_{f,2}(z)$ controller. (B) Plots of controller tracking errors. The switch from 
 $D_{t,1}(z)$ to $D_{f,1}(z)$ is when $e_{t,1}\ge e_{tr,1}=3.88$ and the switch from 
 $D_{t,2}(z)$ to $D_{f,2}(z)$ is when $e_{t,2}\ge e_{tr,2}=4.71$. (C) Duty cycles 
 $d_1$(red) and $d_2$(blue) for the inflation pressure of $F_1$ and $F_2$, respectively. 
 (D) and (E) are plots of $X$ and $Y$ force components for $e_{des,i}=\mu_ie_{tr,i}$, $\mu _i=4$, $i=1,2$  
 on the cylindrical object of a radius $r=2 cm$ and positioned at the location 
 ($X_c, Y_c$) = (0, 15)\,cm.}
\label{fig 4x}
\end{figure}

\emph{Force modulation controller}: Fig.~\ref{fig 4x} shows the results from 
the same setup as in the light touch experiment. The only difference is that at 
the moment of Finger~1 contact detection, we set $R^{ref}_1=y_{@con,1}$ and 
$e_{des,1}=\mu_1  e_{tr,1}=15.52$, where $\mu_1=4$ is a multiplicative factor. At the moment of Finger~2 contact detection, we set $R^{ref}_2=y_{@con,2}$ 
and $e_{des,2}=\mu_2  e_{tr,2}=18.84$, using multiplicative factor $\mu_2=\mu_1=4$. 
Now we can see in Fig.~\ref{fig 4x}D that after the Finger~1 contact, $F_X$ has 
a shift in the value, which is well above the sensor noise level and after the Finger~2 
contact is made, the force $F_X$ value drops. Also, Fig.~\ref{fig 4x}E shows that the force $F_Y$ increases after both Finger~1 and Finger~2 contact 
detections, as expected. 

\emph{In summary, the results illustrated in Fig.~\ref{gentle} and Fig.~\ref{fig 4x} 
confirm our hypothesis of being able to use the controller $D_{f,i}(z)$ to increase 
soft finger contact forces while maintaining the contacts, i.e., to modulate the 
contact force}. 

\begin{figure}[tbp!]
\medskip
\centerline{\includegraphics[scale=0.5]{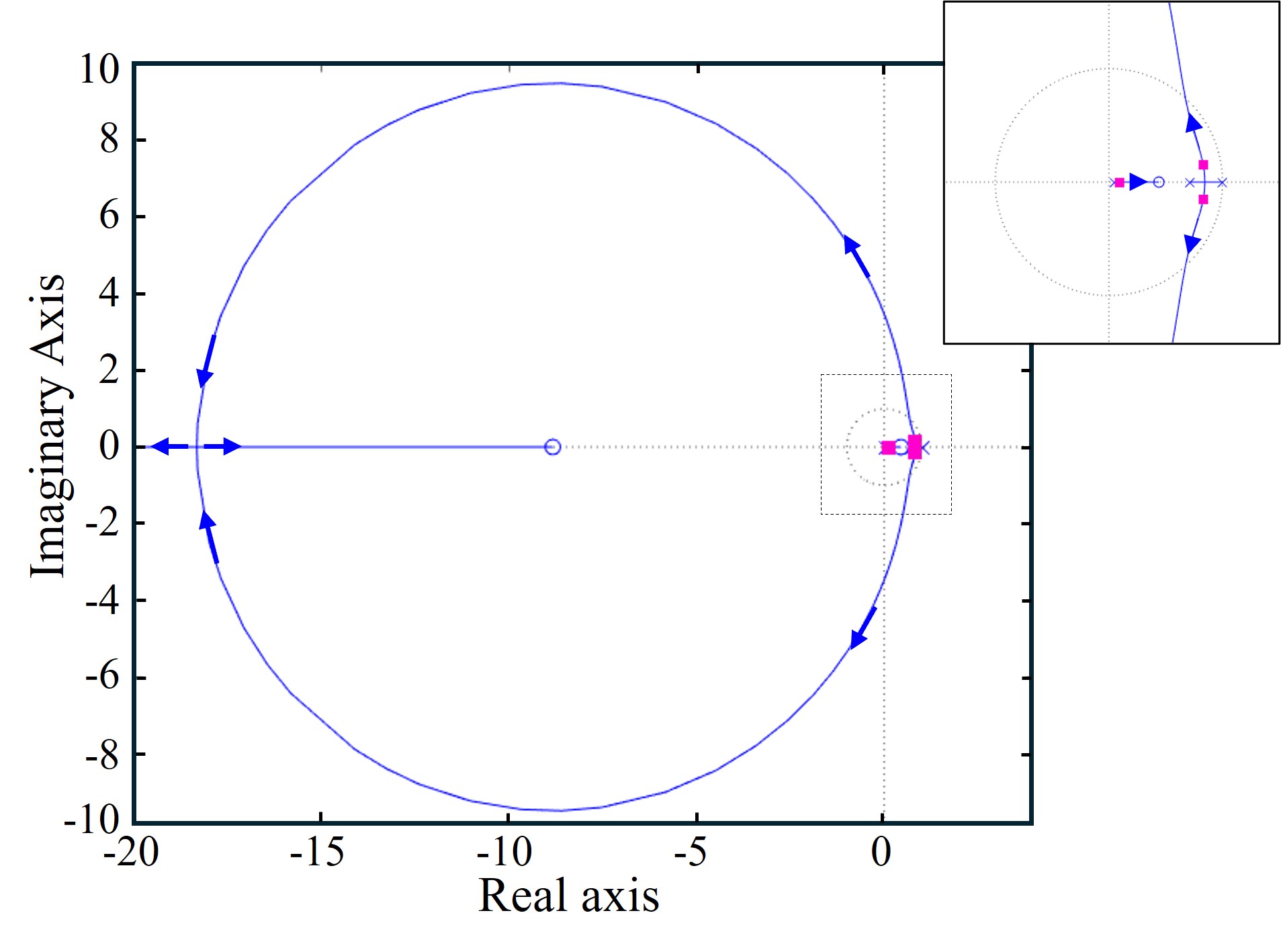}}
\caption{The root locus for the force modulating controller $D_{f,i}(z)$ 
from (\ref{eq:sample3}). The root-locus branches are plotted in solid 
lines with the arrows indicating the branch directions as $K_f$ goes 
from 0 to $\infty$. The symbols 'o' and '$\times$' denote zeros 
and poles of the loop gain $L(z)$. The figure insert shows the part of 
the root locus framed by the dashed line, in which the blue circle represents 
the real zero $z_f$ = 0.44. The magenta squares denote the poles for $K_f=0.21$ 
and they are all inside the unit cycle, therefore, the closed loop system 
is stable.}
\label{fig6}
\end{figure}

\emph{Controller parameters}: The use of the controller $D_{f,i}(z)$ from (\ref{eq:sample3}) 
in the force modulating control loop results in the feedback loop gain
\begin{equation}
L(z)=K_f \frac{z-z_f}{z-1}\frac{B(z)}{A(z)}.
\end{equation}
We used the z-domain root locus from MATLAB's rltool and placed the zero $z_f=0.44$ 
of the feedback loop gain $L(z)$ to obtain the root locus shown in Fig.~\ref{fig6}. The 
root locus branches present paths along which the closed feedback loop poles move as 
$K_f>0$ goes from 0 to $\infty$. Our root locus has three branches. The first 
branch starts ($K_f=0$) from a real pole at $0.05$ and finishes 
$(K_f=\infty)$ at $z_f=0.44$. The other two branches start $(K_f=0)$ from poles at 0.71 and 1, and finish $(K_f=\infty)$ in zero at -8.84 and infinity. In order to have a stable 
feedback control loop, we need to have all closed loop poles within the unit 
circle $|z|=1$. For this reason, we need to take $K_f$ before any of the three  poles 
escapes the circle. After an initial selection of $K_f$ in matlab and its reduction 
and fine tuning on the real system, we finally set $K_f=0.21$.

%% file: 4_ExperimeintResult.tex
\section{Experiments and Results}\label{AA}
We first demonstrate the modulation of contact force using our kinesthetic feedback controller on a fixed object with a single finger. We then extend this capability to a two-finger grasp scenario, illustrating how force modulation can be applied on each finger when an unfixed object is misaligned with respect to the axis symmetry. Finally, we show how our controller adjusts contact forces to modulate pull-out forces. 

\emph{Experiment Setup}: We constructed two pneumatic soft fingers based on the design from the Soft Robotics Toolkit \cite{7862175}. The gripper has a length of 107 mm, with the two fingers mounted 150 mm apart, facing each other in parallel. For bending measurements, we placed three marker stickers on each finger, spaced 27.3 mm apart, using two different colored stickers per finger.

We implemented our controller in Python on a PC, with a sampling time of 100 ms, matching the bending measurement rate from the marker tracking system. The RGB webcam (Logitech C210) was used in conjunction with a marker tracking library in OpenCV.

The air pressure for the grippers was supplied via a pressure control board, adapted from the design provided by the Soft Robotics Toolkit \cite{7862175}. We modified the pressure source to use pressure-regulated wall-compressed air. The board controls the input pressure by adjusting the valve opening through PWM duty cycle, with a control frequency of 90 Hz.

For testing, we 3D-printed a cylindrical object (radius $r =  2$ cm, height 
$= 8.4$ cm) using PLA material. In the contact force modulation tests, we mounted the cylindrical object on a 6 DOF load cell (ATI, Gamma), which sampled data at 2.2 kHz and was fixed to an optical plate. For tests involving grasping an unfixed object, we placed the cylindrical object on an anodized optical plate, ensuring low friction (0.09).

For the pullout force tests, the load cell was mounted on a Franka Emika Research 3 robot arm. At the same time, the cylindrical object was attached to the load cell using cantilevered interconnects to prevent interference from the robot arm with the vision system. Once the two fingers grasped the cylinder object, the robot arm pulled it along the $-Y$ direction at a constant speed (-5 cm/s), measuring the real-time pull-out force.

\subsection{Modulation of Contact Forces With Fixed Object}\label{secVB}
\begin{figure}[b!]
\centerline{\includegraphics[width=0.95\linewidth]{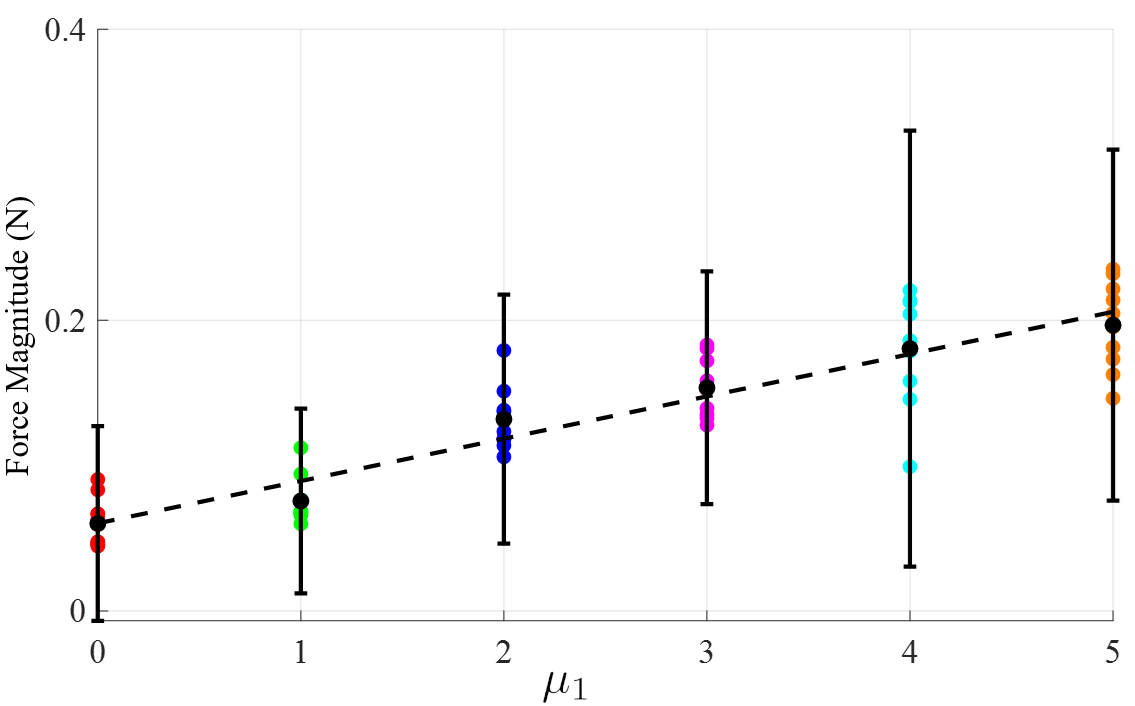}}
\vspace{-\baselineskip}
\caption{Single finger force modulation: the normal force during contact, shown as colored circles, is plotted against the error values of the force controller for the cylinder with radius $r = 2$ cm and position at \( (X_c, Y_c) = (0, 15) \) cm. The mean value for each configuration is shown as a solid black circle, and the best-fit line is depicted as a black dashed line. Gentle single-finger contact can be achieved by setting $e_{des,1}$ = 0 for a resulting normal force of less than $\sim$0.1 N. The maximum force achieved is $\sim$0.25 N with $e_{des,1} = \mu_1 e_{tr,1}=19.4$ when 
the multiplicative factor $\mu_1=5$.}
\label{fig:pinchF_error}
\end{figure}

To test the contact force modulation of our controller (as shown in Fig. \ref{fig 4x}), we positioned the load cell-mounted cylinder at $(X_c=0, Y_c=15)$ cm. 
Fig.~\ref{fig:pinchF_error} shows the plot of Finger 1 (F1)'s contact force magnitudes 
vs. multiplicative factor $\mu_1$, where $e_{des, 1}=\mu_1 e_{tr,1}$.
We found using factor $\mu_i$, $i=1,2$ useful to address the difference in the fingers. We expected to see a linear relation assuming that the soft finger in contact behaves as a loaded spring as described in the pseudo rigid body model\cite{armanini2023soft}. We see that the plot shows the linear relation, although it also shows a slightly larger variation of forces as we applied larger $\mu_1$ values.

\subsection{Asymmetric Grasping of Unfixed Object} 
\begin{figure}[tp!]
\centerline{\includegraphics[width=1\linewidth]{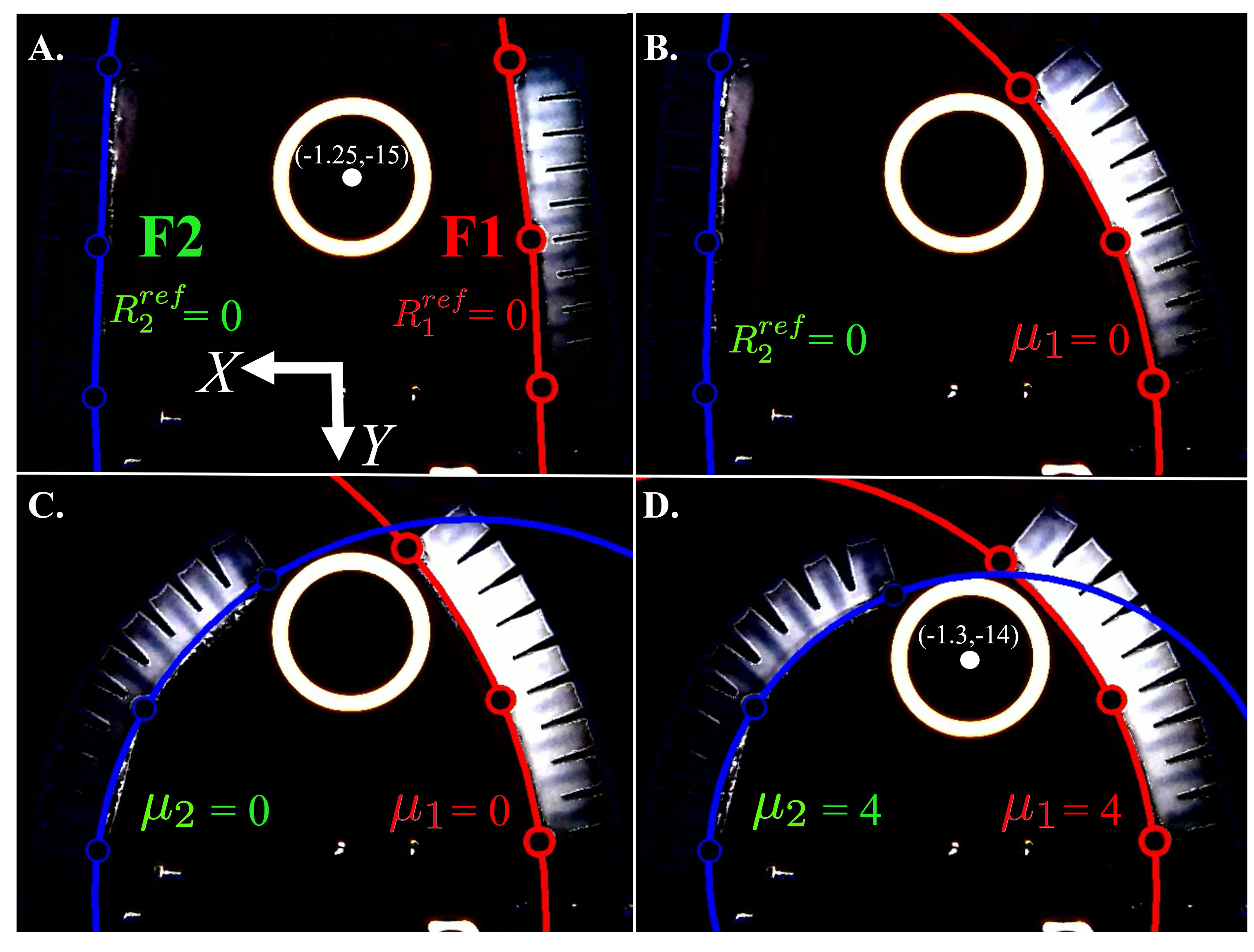}}
\vspace{-\baselineskip}
\caption{Asymmetric grasping: (A) The free-floating cylinder is asymmetrically positioned between F1 and F2. (B) F1 makes contact with a small force $(\mu_1=0)$ such that the cylinder remains stationary. (C) F2 makes 
gentle contact ($\mu_2=0$) while F1 remains in gentle contact ($\mu_1=0$) 
with the cylinder. (D) After the asymmetric gentle grasp is stabilized, the force controllers modulate contact forces ($\mu_1=\mu_2=4$) such that 
the cylinder moves slightly to the right and down from (-1.25, -15) cm to
(-1.3, -14) cm.}
\label{fig:asym}
\end{figure}

Fig.~\ref{fig:asym} depicts 
experiments in which the cylindrical object is positioned at $(X_c=1.25, 
Y_c=15)$ cm. Since $X_c \ne 0$, the left soft finger (F2) has to bend further 
until it reaches contact with the object. 

At the beginning of the experiment, both fingers were controlled with ramp reference tracking controllers. The 
experiments started with $R^{ref}_1=0$ and $R^{ref}_2=0$ resulting in the 
configuration from Fig.~\ref{fig:asym}A. Then we applied a ramp reference 
signal to $R^{ref}_1$, but keeping $R^{ref}_2=0$ until the contact between 
(F1) and the object is detected (see Fig.~\ref{fig:asym}B) when we switched 
from the tracking to the force modulating controller for (F1) with a constant 
$e_{des,1}=\mu_1 e_{tr,1}$ with $\mu_1=0$. After 10 sec, we applied a ramp 
reference signal to $R^{ref}_2$ until the contact between (F2) and the object 
was detected (see Fig.~\ref{fig:asym}C). Then we switched the control of F2 from 
its tracking to the force modulating controller with a constant $e_{des,2}=\mu_2 e_{tr,2}$ 
with $\mu_2=0$. The experiment recorded signals are similar 
to those plotted in Fig.~\ref{gentle}. 
\begin{figure}[b!]
\centerline{\includegraphics[width=0.95\linewidth]{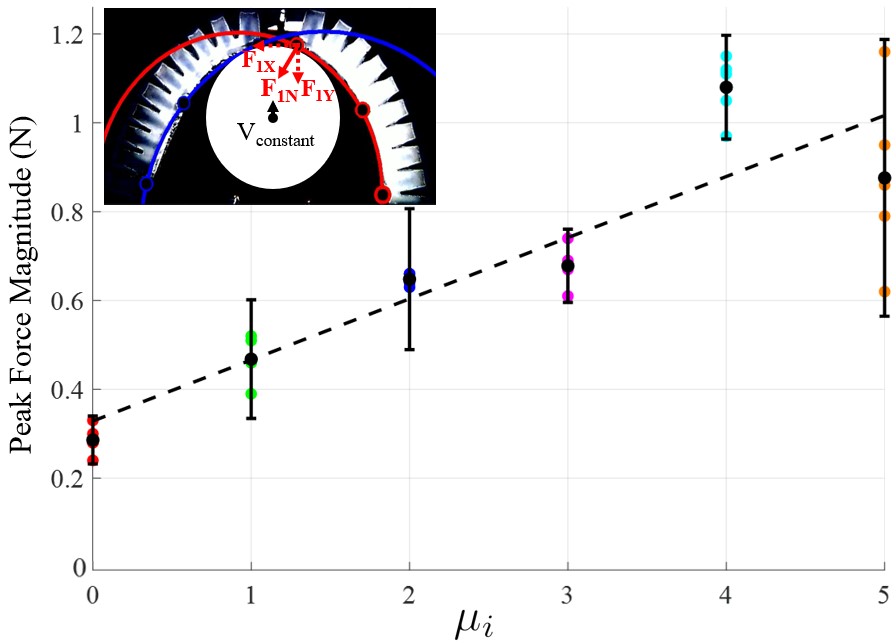}}
\vspace{-\baselineskip}
\caption{The peak pull-out force during power grasping is plotted against the error values of the force controller for the cylinder with radius $r = 2$ cm and position at \( (X_c, Y_c) = (0, 6.25) \) cm. Gentle power grasping can be achieved by setting $e_{des,i}$ = 0 for a resulting pull-out force of $\sim$0.3 N. The maximum pull-out force achieved is approximately $\sim$1.2 N, with $e_{des,i} = \mu_i e_{tr,i}$, $\mu_i=5$, $i=1,2$. The insert illustrates a free-body diagram of the measured force.}
\label{fig:PulloutVSError}
\end{figure}

In this experiment, the cylindrical object was not fixed, but could move 
laterally under the influence of finger contact forces. Our results showed 
that we could achieve contacts between the fingers and object without moving 
the object. Then, once we applied $e_{des, i} = \mu_i e_{tr,i}>0$, $\mu_i=4$, $i=1,2$, i.e., we modulated forces, we were able to move the object. \emph{This result shows that in principle, our approach is capable of manipulating a grasped object.}

\subsection{Modulation of Pull-Out Force}

For the configuration of fingers and the object depicted in the insert of 
Fig.~\ref{fig:PulloutVSError}, we evaluate the pull-out force. We first apply the same 
sequence of references and control as in the previous experiments finishing with 
$\mu_1=\mu_2 \in [0,5]$, $e_{des,i}= \mu_i e_{tr,i}$. After the contact between 
the fingers and the object is stabilized, we initiate the motion of the cylindrical 
object in the $-Y$ direction and record the peak force exerted on the object. 
In this setup, the force sensor and the cylinder are mounted on a 
robot arm (Franka Robotics) which moves the cylinder at a constant 
velocity. 

The plot of peak force magnitude vs. $\mu_1=\mu_2$ values is shown in 
Fig.~\ref{fig:PulloutVSError}. As we can see, the peak force is consistently larger 
than two times the single finger force magnitude from Fig.~\ref{fig:pinchF_error}. 
This is the result of the feedback control coping with keeping the errors $e_{f,i}$, 
$i=1,2$ small, i.e.,  keeping finger bending at the $y_{@con,i}+e_{des,i}$ values under the 
impact of the cylinder motion-induced displacement of the fingers. 
Although the values of used $\mu_1=\mu_2$ do not offer a precise prediction 
of the peak pull-out force magnitude, 
they modulate the peak force through a roughly linear relation shown 
in the Fig.~\ref{fig:PulloutVSError} plot.\\

%% file: 5_Discussion.tex
\section{Conclusion}\label{AB}

Our framework offers feedback control of the contact force of soft gripper fingers using kinesthetic (i.e., bending) sensing without the need for direct tactile or pressure sensors.  The experiments show that the finger contact forces and their pull-out force are 
in a roughly linear relation with our control parameter, i.e., desired reference error. 
They also demonstrate that with 
our approach the controller can establish gentle contact with both fingers, irrespective 
of whether the object is symmetrically or asymmetrically positioned, and fixed or not. Furthermore, we show in Fig.~\ref{fig:asym} that we can modulate the force to move a grasped object.

Our approach should not only reduce the costs, but also enhance 
the reliability and ease of implementation of soft grippers. The sole requirement 
is the incorporation of any curvature or bending sensor. We used vision-based sensing, 
but any off-the-shelf stretchable bending sensors are also applicable. 
Potential applications of our approach include handling delicate objects, such as food, or 
harvesting fruits, where gentle force modulation is crucial. 

In future work, we plan to implement our framework on a physical robotic arm equipped 
with soft dual-fingers, similar to \cite{li2022resonant}. Using reference signals, and 
feedback generated control and error variables, we will explore the use of machine learning 
for in-hand manipulation \cite{10146043}, and more adaptive and robust contact detection and grasping.